# Statistical Approach for Selecting Elite Ants

**Raghavendra G. S.**
**Department of Computer Science,**
**BITS Pilani - K. K. Birla Goa Campus, Zuarinagar, Goa, India**

**Prasanna Kumar N**
**Department of Mathematics,**
**BITS Pilani - K. K. Birla Goa Campus, Zuarinagar, Goa, India**

**ABSTRACT:** Applications of ACO algorithms to obtain better solutions for combinatorial optimization problems have become very popular in recent years. In ACO algorithms, group of agents repeatedly perform well defined actions and collaborate with other ants in order to accomplish the defined task. In this paper, we introduce new mechanisms for selecting the Elite ants dynamically based on simple statistical tools. We also investigate the performance of newly proposed mechanisms.

## Introduction

Nature has been a source of inspiration for various computing paradigm. One such computing paradigm is Ant Colony Optimization which is inspired by foraging behavior of ants. The foraging behavior of ants has fascinated many researchers, which has lead to the development of various ant computational models. In search of food, ants leave their nest and move towards the food source in a random direction. On their journey towards the food source, they leave behind a chemical substance called pheromone trials. The laid pheromone trial will help the ants to trace their way back to nests and guide the other fellow ants to make a journey towards the food source. Thus, pheromone trial will act as an indirect communication medium through which the individual ants share their journey experiences. The sharing experience mechanism will help the ants in establishing the shortest path between the food source and the nest. Ants are able to establish shortest path because the amount of pheromone concentration on shortest path will be higher compared to the other paths making it more favorable path to follow. The above observed phenomena have lead to the development of Ant algorithms [DMC96]. The new computing paradigm ACO has a feature of positive feedback, distributed computation and use of constructive greedy heuristic approach. The positive feedback in the form of reinforcement guides the ant in search process. The search process is distributive in nature due to involvement of multiple ants. The distributive computation avoids the premature convergence by thoroughly exploiting the search space. The greedy heuristic mechanism helps in finding the acceptable solution. A close observation reveals that food hunting behavior can be used to attack the combinatorial problems which are non polynomial in nature by constructing the artificial ant models. In literature, many variants of ant algorithms have been proposed and each algorithm improvises the earlier versions [GD97], [BRD04], [RP08]. These improvised algorithms try to strike the balance between exploration and exploitation. An Ant which exploits the search space around the optimally best solution may not get globally best solution. Similarly, exploring the search space will get the globally best solution, but need more time to converge. Therefore it is necessary to strike the balance between exploration and exploitation for better performance in terms of quality of solution found and the optimal time needed to converge. However, general outline of the ACO algorithm is as follows:

Set parameters, initialize pheromone trials
While termination condition not met do
Construct Ant Solutions
Apply Local Search (optional)
Update Pheromone trials
End while

One of the standard benchmark problems available to evaluate the ACO algorithm is Travelling Salesman Problem (TSP), where ants will solve the Hamiltonian problem. A standard framework model can be defined for ACO algorithms consisting of mainly two phases; namely,

- *Solution construction phase* - Each ant is assigned a task for completion.
- *Update phase* - Each ant shares the experience of completing the task in the form of pheromone trial updation.

In TSP, $m$ ants will make a tour of $n$ cities. After the completion of tour, they will update the corresponding paths proportional to the quality of solution found by them. The updated paths provide valuable knowledge for the ants in the search process. The pheromone updation can be

- *Global best updation* - The best path from the start of the execution is updated.
- *Iteration best updation* - iteration best tour is updated.

The performance of iteration best updation is better than the global best updation due to better search space exploitation. In addition, several options are available for updation like number of ants used for updation and best path updation. The updation with respect to number of ants used can be

- *Communism* - where all the ants are allowed to update the path
- *Elitism* - only the best ant is allowed to update the path.
  The updation with respect to best tour can be
- *Global best tour* - pheromone updation is same for all the links of the path based on the overall quality of solution found and done at the end of tour.
- *Local best tour* - The trial reinforcement is done for individual links along the path and it is proportional to the contribution the individual links make to the final solution. The updation is done immediately after the completion of corresponding link tour.

The complete execution of solution construction phase and updation phase constitutes the single iteration of the algorithm. In order to assess the performance, algorithm is executed for predefined number of iterations or until the solution converges.

## 1. Basic Ant Colony Algorithms

The Ant algorithms have been successfully applied to various benchmark problems like Traveling Salesman Problem (TSP), Job-Shop Scheduling (JSP), Vehicle Routing Problem (VRP), Graph Coloring Problem (GRP) and Quadratic Assignment Problem (QAP) and have been extended to continuous search domain also. Based on the literature survey, research work related to ACO can be classified into following categories: Devising new strategies for pheromone updation [JSF04], Reward – Penalty approaches, Dynamic parameter adjustment [LSY08], [MMH08], Hybridization of ant algorithms [Blu07], Proof for convergence [SD02], [HF06], [PR11a] and Applying the ant algorithms to multi-disciplinary fields. In this section, we provide brief overview of some of the important variants of ant algorithms.

## 1.1. Ant System (AS)

The first ant algorithm proposed was AS [DMC96]. The algorithm works as follows: During the solution construction phase, ants are randomly placed and they are asked to complete the tour. A transition probability function $p_{ij}$ is defined for the ants in order to make decision to select the next city to be visited. Suppose the ant is in city $i$ and it needs to make a move to the next city $j$ and the probability of selecting city $j$ is given by the equation:

$$p_{ij}(t) = \frac{[\tau_{ij}]^\alpha [\eta_{ij}]^\beta}{\sum_{j \in N_i} [\tau_{ij}]^\alpha [\eta_{ij}]^\beta} \qquad (1)$$

where $\tau_{ij}$ is the pheromone strength on link $ij$, $\eta_{ij}$ is the visibility of link $ij$, $\alpha$ and $\beta$ are parameters, which control the importance of pheromone strength and visibility and $N_i$ is the feasible neighborhood of the ant when being at city i. The algorithm follows the communism approach and the pheromone updation is given by the expression:

$$\tau_{ij} = \rho.\tau_{ij} + \Delta\tau_{ij}$$
$$\Delta\tau_{ij} = \sum_{k=1}^{m} \Delta\tau_{ij}^k$$

where $\rho$ is the pheromone persistent factor and $\Delta\tau_{ij}^k$ is computed according to the equation:

$$\Delta\tau_{ij}^k = \begin{cases} Q/L_k & \text{if (i, j) } \epsilon \text{ } k\text{-th ant's tour list} \\ 0 & \text{otherwise} \end{cases}$$

where $Q$ is the algorithmic constant and $L_k$ is the tour length of $k^{th}$ ant.

The paper [DMC96] discusses the three variants of ant algorithm namely ant-cycle, ant- density and ant-quantity and each of them differs in the way the pheromone is updated. An extension to ant algorithm called "Elitist strategy" has been proposed, where best-so-far tours will be reinforced again after the standard reinforcement. The additional reinforcement is given by $e \cdot Q/L^*$ where $e$ is the number of elitist ants and $L^*$ is the length of best tour. The disadvantage of the ant system is that its performance suffers for larger problem size and search stagnation occurs much earlier without proper exploitation.

## 1.2. Ant Colony System (ACS)

ACS is a modified version of AS [GD95]. In solution construction, ants select the next city to be visited based on pseudo-random-propositional rule. Suppose the ant is in city $r$ and would probabilistically select the city $s$ according to the equation:

$$s = \begin{cases} \underset{u \, \epsilon \, J_k(r)}{arg \; max} \quad [\tau \, (r,u)] \, . \, [\, \eta \, (r,u)]^{\beta} & \text{if } \; q \leq q_0 \, (\text{Exploitation}) \\ S & \text{otherwise (Exploration)} \end{cases} \quad (2)$$

where q is a random number uniformly distributed between [0, 1], $q_0$ is a parameter in the range $(0 < q_0 < 1)$ and S is a random variable selected according to the probability distribution given in Eq (1). The pheromone updation phase consists of local updation and global updation. The global updation is given by the equation:

$$\tau \, (r, s) \leftarrow \; (1 - \alpha) \cdot \tau \, (r, s) + \tau \cdot \Delta\tau \, (r, s)$$

and

$$\Delta\tau \, (r, s) \; = \begin{cases} (L_{gb})^{-1} & \text{if } (r, s) \, \epsilon \text{ global best tour} \\ 0 & \text{otherwise} \end{cases}$$

where $0 < \alpha < 1$ is the pheromone decay parameter, $L_{gb}$ is the global best tour. Similarly, local updation is done by all the ants and is given by the equation:

$$\tau \, (r, s) \leftarrow \; (1 - \rho) \cdot \tau \, (r, s) + \rho \cdot \Delta\tau \, (r, s) \quad (3)$$

where $0 < \rho < 1$ is a pheromone reinforcement parameter and $\Delta\tau \, (r, s)$ is set to $\tau_0$. The proposed algorithm was able to find the good optimal tour in lesser number of iteration, but amount of work done by ants is more compared to other variants of ant algorithms.

## 1.3. Rank-based Ant System (AR)

The basic idea of elitist ants has been incorporated into the AR [BHS99]. In addition to that, each ant is assigned rank according to its performances and the assigned rank is used for the pheromone updation purpose. The solution construction phase is same as the AS. In pheromone updation phase, updation is done twice for the best selected tours ($\sigma$) and is given by the equations.

$$\tau_{ij} = \rho.\tau_{ij} + \Delta \tau_{ij} + \Delta \tau_{ij}^{*} \qquad (5)$$

where

$$\Delta \tau_{ij} = \sum_{\mu=1}^{\sigma-1} \Delta \tau_{ij}^{\mu}$$

$$\Delta \tau_{ij}^{\mu} = \begin{cases} (\sigma\text{-}\mu).Q/L_{\mu} & \text{if the } \mu\text{-th best ant travel on edge(i, j)} \\ 0 & \text{otherwise} \end{cases}$$

and

$$\Delta \tau_{ij}^{*} = \begin{cases} (\sigma\text{-}\mu).Q/L^{*} & \text{if edge(i, j) is a part of the best solution} \\ 0 & \text{otherwise} \end{cases}$$

Here $\Delta \tau_{ij}^{*}$ specifies the amount of pheromone increases due to elitist ants, $\mu$ specifies the ranking index and $\sigma$ is the number of elitist ants.

In first updation, trial contribution by the ant is based on its tour performance and in second updation trial contribution is weighted against the performance and accordingly done. If a given ant has found the better solution, then its ranking will be better and hence better trial contribution. One of the disadvantages of the rank system is, its inability to find the good optimum result. Even then the result obtained was nearer to the optimum solution.

## 1.4. Max-Min Ant System (MMAS)

MMAS is a modified version of elitist ant system [SH00]. The solution construction phase is same as the AS but it differs in the pheromone updation phase in two aspects. The first difference is the elite approach used for pheromone updation with the intention to exploit the best solution. The second difference is pheromone strength $\tau_{ij}$ on all the edges will be in the range specified by $\tau_{min}$ and $\tau_{max}$ where $\tau_{min} < \tau_{ij} < \tau_{max}$. This will help the algorithm to avoid the early search stagnation. The updation of best path can be done with either global best or iteration best tour. This system also involves the branching factor, a technique to determine whether algorithm has converged or not. The branching factor is given by the equation:

$$\tau > \tau_{il}^{\min} + \lambda.(\tau_{ij}^{\max} - \tau_{ij}^{\min})  \tag{5}$$

The branching factor is the number of outgoing edges satisfying the Eq (5). The Average branching factor is computed by considering all the edges. If $\tau$ value is approximately 1, then there exists only one edge for exiting the node, indicating that ants have found the better path. In order to explore the new tours, new mechanisms called smoothing of trials have been proposed. The smoothing mechanism will adjust the trial intensity on all edges by $\tau_{\max} - \tau_{ij}$ factor, there by facilitating the search in an unexplored region. Although MMAS is capable of finding the good optimal solution, it takes more time to converge.

## 2. Motivation

In this paper, we suggest elite selection mechanism for Elitist Ants (EA) and Rank based ants (RA). The proposed variant is inspired by the EA and RA algorithms, where pre-defined number of ants is selected for the second time reinforcement. The paths that are updated in second time reinforcement are called elite paths. The motivation for proposing the algorithms comes from the following observation. The First observation is that the papers due to [DMC96], [BHS99] do not discuss the criteria for selecting the elitist ants and all the ants are treated as elitist. In fact, the program execution reveals that if smaller number of ants were selected as elitist, the exploitation of search area will be restricted near the good solutions and this may not contribute much to the final quality of solution. Similarly, if large number of ants were selected as elitist then most of the paths will get additional reinforcement leading to better exploration of search space. Although there will be improvement in quality of solution, but in due process there will be reinforcement for some paths that may not contribute to the final solution. This necessitates that optimum number of ants needs to be selected for second time reinforcement.

The second observation reveals that at the early stage of program execution, ants gain information about the search space through exploration. It can be argued that size of the search space will be large at the initial stage and will get reduced at the later stage of execution due to knowledge gained by the ants about the search space in exploration phase. The additional reinforcement of appropriate paths in exploration stage should help ants in exploitation stage for better solution. The algorithm in execution should update comparatively larger number of elite paths during exploration phase and fewer numbers of paths upon transition to exploitation phase. It is interesting to observe that overall search region is going to be dynamic in nature and the dynamicity can be introduced by selecting appropriate number of elite ants. In this paper we will suggest the mechanism for selecting the elite ants purely based on the ant's performance using statistical functions.

## 3. Machine Learning

This section will provide the brief outline of the machine learning. Machine Learning (ML) [Eth05] is a discipline of computing field concerned with training the machines to perform certain tasks. A machine learns to perform the task by gaining knowledge and by remembering the past experiences. The acquired knowledge and experience will be used to make the decisions that are necessary to solve the task. The underlying basis for learning mechanism is the statistical data collected from the observation and the data that will evolve in the future. The collected data will be used to train the machine, so that it can take appropriate decision for solving the task and the evolved data is used to adjust the decision making attitude of the machines in order to improve the accuracy and performance. The ultimate goal of the ML is to mimic the human intelligence in machines. Machine learning is an interdisciplinary field borrowed the idea from other fields like statistics, pattern recognition, artificial intelligence, adaptive control theory, evolutionary models etc. A machine can be trained to solve some of the tasks like:

- *Classification* - The process in which machine will act as a classifier and assign the data to the groups/classes they belong.
- *Prediction* - The mechanism is similar to classification, where a machine is going to predict the class of the incoming data based on past experience and knowledge.
- *Rule Generation* - The mechanism is all about generating the rules by looking into the relation that exist between the data

- *Clustering* - The process of grouping the data by looking at the similarities present in the data.

   The learning mechanism can be:

1 *Supervised learning* - A machine have the knowledge about the number of classes and characteristic features of each class. Initially, machine will be trained with few samples of data to perform the task. Classification and prediction tasks fall under this category of learning.

2 *Unsupervised learning* - In this learning mechanism, machine have no knowledge about the number of classes and characteristic features about the classes. In fact, machine learns by performing task. Clustering is an example that falls under this category of learning.

3 *Reinforcement learning* - A learning mechanism that specifies the action need to be taken for each observations and reward the action in the form of feedback that guides the learning process.

   The ML methods has lot of application in real life and some of the them like analyzing customer buying pattern in supermarket, face recognition, stock market prediction etc have been commercially deployed.

## 4. The Selection Mechanism

In this section, we will discuss the selection mechanism of dynamic ants. The selection process can be treated as a two class classification problem, where we need to design the classifier that will automatically classify the ants depending upon their performances. The classifier will place performing ants into one class and non-performing ants into another class. It can be observed that ants have to remember the class they belong in addition to tour length. The statistical tools will provide the boundary that separates the two classes. The following statistical tools have been used in the design of classifier:

- *Mid - Range Tour Selection (MRTS):* MRTS computes the average of a tour by considering best tour length and worst tour length in a given iteration and is given by the equation:

$$MRTS = (Best\ Tour\ Length + Worst\ Tour\ Length)/2 \qquad (6)$$

- *Mean Tour Selection (MTS):* MTS computes the mean of the tour by considering all the tour lengths and is given by the equation:

$$Mean = 1/n\sum_{i=1}^{n}TL_i \qquad (7)$$

- *Median Tour Selection (MeTS):* Computes the median by arranging all the tour lengths $TL_i$ in the increasing order i.e., $TL_1 \leq TL_2 \leq \cdots \leq TL_n$ and then select $TL_{n/2}$ as the median value.

The computed values of mid-range, mean and median in each iteration will provide the boundary for the classes.

## 5. The Proposed algorithms and its extension

In this section, we will discuss the incorporation of selection mechanism into two versions of ant algorithms namely EA and RA. The selection mechanism will place all performing ants into one class and non-performing ants into another class based on statistical function. Suppose mean function is used for classification. Then tour performances lesser than mean will be in performing class and rest of the ants will be in non-performing class. In fact performing ants will get chance for additional reinforcement according to the algorithmic specification. The algorithmic specification for EA is given by the expression *e.Q/L* and for RA; it is same as Eq (4). It should be noted that number of elite ants varies across the iterations compared to traditional EA and RA, where the number of elite ants were fixed as a part of parameter settings. Since the number of selected ants varies across the iteration, we will name them as Dynamic Ants (DA). The proposed algorithms here onwards are named as Dynamic Elitist Ants (DEA) and Dynamic Rank Ants (DRA) due to incorporation of EA and RA in DA.

We further extend the dynamic elitism by incorporating punishment feature, where non performing ants will be punished by removing the pheromone on the path they have travelled. The basic purpose of introducing the punishment mechanism is to favor exploitation process by restricting the search in promising area of the search space. The quantity of pheromone to be removed will be specified by the algorithmic specification. The algorithmic specification for EA specifies to decrease the quantity of pheromone trial proportional to the quality of solution found on non elite paths and in case of RA, all the non elite paths are weighted according to their performances and accordingly decreased. The generic punishment feature for EA and RA is given by the equation:

$$\tau_{ij} = \tau_{ij} - \Delta\tau_{ij}^{*}$$

where
$$\Delta \tau_{ij}^{*} = \sum_{k=1}^{l} \Delta \tau_{ij}^{k}$$

Also $\Delta \tau_{ij}^{k}$ in case of EA is given by the equation:

$$\Delta \tau_{ij}^{k} = \begin{cases} e.Q^{*}/L_{k} & \text{if (i, j) є k-th ant's non performing tour list} \\ 0 & \text{otherwise} \end{cases}$$

and the $\Delta \tau_{ij}^{k}$ in case of RA is given by the equation:

$$\Delta \tau_{ij}^{k} = \begin{cases} Q^{*}.(m-k)/L_{k} & \text{if (i, j) є k-th ant's non performing tour list} \\ 0 & \text{otherwise} \end{cases}$$

where $Q^*$ is the algorithmic constant. If $e$ represents the number of elite ants in a particular iteration, then $L$ represents the non-elite ants given by the expression $L = m - e$. It should be noted that number of performing ants and non performing ants vary across the iteration and it will be determined by ant's performance in previous iteration.

## 6. Experimental Studies

In order to demonstrate the superiority of the proposed algorithm, we compared our results with MMAS, a best variant among the existing ant algorithm available in the literature. To have a better assessment of the proposed algorithm, parameters α, β were varied from 1 to 5 and ρ was varied from 0.7 to 1.0. The proposed algorithms were executed for 10 times independently by considering some of the datasets available in the TSPLIB. The maximum number of iteration was set to 100000. The Table 1 provides the comparative results for Dynamic Elitist Ants (DEA) and Dynamic Rank Ants (DRA) algorithm incorporated with statistical tool (Mean, Median and Mid-Range tour selection). The proposed algorithms are compared with MMAS+IB+PTS for best solution, average solution and the percentage of deviation from the optimal solution. The average solution was computed using the best solutions of last 50 iterations. The proposed algorithms were able to find the better solutions for most of the datasets. The Table 1 shows that DEA incorporated with MTS provides best solution for att48 dataset with deviation of 0.04% and with incorporation of MeTS provides best solution for st70 dataset with deviation of 0.05%. The incorporation of MTS in DRA provides best solution for Kroa100, lin318 with observed deviation of 0.01% and with MeTS incorporation, best results were obtained for Kroa100 and Kroa200 with 0% deviation.

In general, it can be concluded that DEA provides better results for smaller dimension problems and DRA provides better results for higher dimension problems. The solutions provided by MRTS deviate more from the optimal solution compared to other tour selection mechanism for both DEA and DRA algorithms and it can be attributed to updation of not so promising paths. The algorithmic simulation suggests that median function takes lesser number of iterations to find optimal solution than the mean function. Another interesting observation is that, average solutions of proposed algorithms for most of the datasets have larger deviations from the optimal solutions indicating lack of focus to concentrate on promising region of the search spaces.

**Table 1: Comparison of performance of Dynamic Ants with MMAS**

| Datasets | Algorithms | Best (Std Dev) | Average (Std Dev) |
|---|---|---|---|
| bays29 | MMAS+IB+PTS | 2022.1 (0.1%) | 2025.3 (0.26%) |
| | DEAMR | 2039.4 (0.96%) | 2055.6 (1.76%) |
| | DEAM | 2034.7 (0.72%) | 2046.9 (1.33%) |
| | DEAMed | **2021.7 (0.08%)** | 2047.3 (1.35%) |
| | DRAMR | 2044.4 (1.20%) | 2059.5 (1.95%) |
| | DRAM | 2040.8 (1.02%) | 2055.9 (1.77%) |
| | DRAMed | 2042.2 (1.09%) | 2057.8 (1.87%) |
| att48 | MMAS+IB+PTS | 10634.4 (0.06%) | 10640.8 (0.12%) |
| | DEAMR | 10821.8 (1.82%) | 10892.4 (2.48%) |
| | DEAM | **10632.8 (0.04%)** | 10695.1 (0.63%) |
| | DEAMed | 10638.2 (0.09%) | 10702.7 (0.7%) |
| | DRAMR | 10768.9 (1.32%) | 10844.7 (2.03%) |
| | DRAM | 10656.8 (0.27%) | 10690.5 (0.58%) |
| | DRAMed | 10643.8 (0.14%) | 10697.4 (0.65%) |
| eil51 | MMAS+IB+PTS | 426.2 (0.04%) | 427.8 (0.43%) |
| | DEAMR | 438.4 (2.42%) | 448.3 (4.74%) |
| | DEAM | 426.5 (0.11%) | 442.6 (3.41%) |
| | DEAMed | 426.8 (0.18%) | 433.2 (1.21%) |
| | DRAMR | 436.2 (1.91%) | 444.9 (3.94%) |
| | DRAM) | 434.6 (1.54%) | 442.7 (3.43% |
| | DRAMed | 431.6 (0.84%) | 439.7 (2.73%) |

| Datasets | Algorithms | Best (Std Dev) | Average (Std Dev) |
|---|---|---|---|
| **st70** | MMAS+IB+PTS | 675.5 (0.07%) | 680.3 (0.78%) |
| | DEAMR | 712.7 (5.58%) | 730.4 (8.20%) |
| | DEAM | 678.5 (0.51%) | 710.1 (5.2%) |
| | DEAMed | **675.4 (0.05%)** | 698.3 (3.45%) |
| | DRAMR | 685.3 (1.52%) | 710.7 (5.28%) |
| | DRAM | 678.5 (0.51%) | 689.4 (2.13%) |
| | DRAMed | 677.2 (0.32%) | 686.4 (1.68%) |
| **eil76** | MMAS+IB+PTS | 538.5 (0.09%) | 539.9 (0.35%) |
| | DEAMR | 552.6 (2.71%) | 582.4 (8.25%) |
| | DEAM | 545.7 (1.43%) | 551.4 (2.49%) |
| | DEAMed | 547.6 (1.78%) | 561.2 (4.31%) |
| | DRAMR | 554.3 (3.02%) | 575.7 (7%) |
| | DRAM | 542.4 (0.81%) | 549.8 (2.19%) |
| | DRAMed | 540.3 (0.42%) | 548.7 (1.98%) |
| **Kroa100** | MMAS+IB+PTS | 21285.4 (0.01%) | 21336.9 (0.26%) |
| | DEAMR | 21890.8 (3.28%) | 22140.7 (4.03%) |
| | DEAM | 21540.6 (1.21%) | 21598.6 (1.48%) |
| | DEAMed | 21385.7 (0.48%) | 21456.9 (0.82%) |
| | DRAMR | 21780.4 (2.34%) | 21930.7 (3.04%) |
| | DRAM | **21284.6 (0.01%)** | 21436.3 (0.72%) |
| | DRAMed | **21283.8 (0%)** | 21385.6 (0.48%) |
| **kroa200** | MMAS+IB+PTS | 29372.2 (0.01%) | 29385.8 (0.06%) |
| | DEAMR | 31468.9 (7.15%) | 31790.5 (8.24%) |
| | DEAM | 29640.6 (0.92%) | 29689.9 (1.09%) |
| | DEAMed | 29536.7 (0.57%) | 29590.3 (0.75%) |
| | DRAMR | 29840.6 (1.60%) | 29994.8 (2.13%) |
| | DRAM | 29390.9 (0.07%) | 29569.4 (0.68%) |
| | DRAMed | **29370.3 (0%)** | 29480.8 (0.38%) |
| **lin318** | MMAS+IB+PTS | 42035.7 (0.01%) | 42055.8 (0.06%) |
| | DEAMR | 44896.5 (6.82%) | 45218.7 (7.58%) |
| | DEAM | 43220.6 (2.83%) | 43312.7 (3.05%) |
| | DEAMed | 42870.4 (2%) | 42910.3 (2.09%) |
| | DRAMR | 43723.6 (4.03%) | 43890.4 (4.42%) |
| | DRAM | **42034.5 (0.01%)** | 42392.3 (0.86%) |
| | DRAMed | 42137.2 (0.25%) | 42297.8 (0.63%) |

The Table 2 shows the incorporation of punishment mechanism to dynamic ants. On comparison with Table 1, it can be observed that punishment mechanism improvises the solution obtained by DA and also obtains the best optimal solution for some of the datasets. The quality of solution found by MRTS was inferior and exhibits larger deviation from the optimal solution. The punishment mechanism improvises the average solution for most of the datasets under consideration, demonstrating the algorithms ability in restricting the search in promising area of search space.

**Table 2: Comparison of performance of Punished Dynamic Ants with MMAS**

| Datasets | Algorithms | Best (Std Dev) | Average (Std Dev) |
|---|---|---|---|
| **bays29** | MMAS+IB+PTS | 2022.1 (0.1%) | 2025.3 (0.26%) |
| | DEAMR_pun | 2053.5 (1.65%) | 2082.1 (3.07%) |
| | DEAM_pun | 2022.4 (0.11%) | 2027.8 (0.38%) |
| | DEAMed_pun | **2021.5 (0.07%)** | **2023.6 (0.17%)** |
| | DRAMR_pun | 2052.8 (1.62%) | 2074.4 (2.69%) |
| | DRAM_pun | 2038.9 (0.93%) | 2044.9 (1.23%) |
| | DRAMed_pun | 2034.7 (0.72%) | 2044.6 (1.21%) |
| **att48** | MMAS+IB+PTS | 10634.4 (0.06%) | 10640.8 (0.12%) |
| | DEAMR_pun | 10730.5 (0.96%) | 10834.4 (1.94%) |
| | DEAM_pun | **10632.2 (0.03%)** | 10645.2 (0.16%) |
| | DEAMed_pun | **10630.4 (0.02%)** | **10638.8 (0.1%)** |
| | DRAMR_pun | 10656.7 (0.27%) | 10712.4 (0.79%) |
| | DRAM_pun | 10643.2 (0.14%) | 10668.8 (0.38%) |
| | DRAMed_pun | 10645.9 (0.16%) | 10674.6 (0.43%) |
| **eil51** | MMAS+IB+PTS | 426.2 (0.04%) | 427.8 (0.43%) |
| | DEAMR_pun | 440.6 (2.94%) | 455.3 (6.37%) |
| | DEAM_pun | 426.4 (0.09%) | 434.3 (1.47%) |
| | DEAMed_pun | **426.1 (0.02%)** | **427.4 (0.32%)** |
| | DRAMR_pun | 440.5 (2.92%) | 452.3 (5.67%) |
| | DRAM_pun | 432.7 (1.09%) | 438.9 (3.02%) |
| | DRAMed_pun | 430.8 (0.65%) | 437.3 (2.65%) |
| **st70** | MMAS+IB+PTS | 675.5(0.07%) | 680.3(0.78%) |
| | DEAMR_pun | 698.4 (3.46%) | 720.3 (6.71%) |
| | DEAM_pun | **675.3 (0.04%)** | **678.6 (0.53%)** |
| | DEAMed_pun | 680.8 (0.85%) | 685.3 (1.52%) |
| | DRAMR_pun | 691.5 (2.44%) | 715.6 (6.01%) |
| | DRAM_pun | 676.5 (0.22%) | 684.5 (1.40%) |
| | DRAMed_pun | **675.4 (0.05%)** | 682.7 (1.14%) |

| Datasets | Algorithms | Best (Std Dev) | Average (Std Dev) |
|---|---|---|---|
| **eil76** | MMAS+IB+PTS | 538.5 (0.09%) | 539.9 (0.35%) |
| | DEAMR_pun | 561.4 (4.34%) | 584.3 (8.60%) |
| | DEAM_pun | 541.4 (0.63%) | 548.4 (1.93%) |
| | DEAMed_pun | 545.6 (1.41%) | 555.4 (3.23%) |
| | DRAMR_pun | 548.5 (2%) | 564.3 (4.88%) |
| | DRAM_pun | 541.4 (0.63%) | 545.6 (1.41%) |
| | DRAMed_pun | 538.8 (0.14%) | 543.7 (1.05%) |
| **Kroa100** | MMAS+IB+PTS | 21285.4 (0.01%) | 21336.9 (0.26%) |
| | DEAMR_pun | 21780.4 (2.34%) | 21867.3 (2.75%) |
| | DEAM_pun | 21321.6 (0.18%) | 21375.1 (0.43%) |
| | DEAMed_pun | 21330.7 (0.22%) | 21367.8 (0.4%) |
| | DRAMR_pun | 21610.6 (1.54%) | 21688.7 (1.91%) |
| | DRAM_pun | 21286.3 (0.02%) | **21295.7 (0.06%)** |
| | DRAMed_pun | **21282.7 (0%)** | **21288.4 (0.03%)** |
| **kroa200** | MMAS+IB+PTS | 29372.2 (0.01%) | 29385.8 (0.06%) |
| | DEAMR_pun | 30850.7 (5.04%) | 31224.7 (6.32%) |
| | DEAM_pun | 29540.8 (0.58%) | 29588.4 (0.75%) |
| | DEAMed_pun | 29444.8 (0.26%) | 29489.2 (0.41%) |
| | DRAMR_pun | 29664.8 (1.01%) | 29710.7 (1.16%) |
| | DRAM_pun | 29446.5 (0.26%) | 29486.5 (0.40%) |
| | DRAMed_pun | **29368.5 (0.0%)** | **29380.8 (0.04%)** |
| **lin318** | MMAS+IB+PTS | 42035.7 (0.01%) | 42055.8 (0.06%) |
| | DEAMR_pun | 44219.6 (5.12%) | 44870.4 (6.76%) |
| | DEAM_pun | 42780.5 (1.78%) | 42932.7 (2.15%) |
| | DEAMed_pun | 42540.8 (1.21%) | 42624.3 (1.41%) |
| | DRAMR_pun | 43879.5 (4.40%) | 43964.3 (4.60%) |
| | DRAM_pun | **42033.2 (0%)** | **42042.9 (0.03%)** |
| | DRAMed_pun | 42045.7 (0.03%) | 42094.3 (0.15%) |

The Table 3 provides the details of average number of ants selected for second time updation and the total number of ants present in the system, when optimal solution is obtained. It can be observed that more than 70-80% ants in the system are selected for additional reinforcement in MRTS mechanism and 60-70% of ants in case of MTS and MeTS. The selection of larger number of ants for additional reinforcement in MRTS indicates the algorithm inability for not obtaining the optimal solution.

**Table 3: Average number of ants selected for second time updation for Dynamic Ants**

| Data Sets | DEAMR | DEAM | DEA Med | DRAMR | DRAM | DRA Med |
|---|---|---|---|---|---|---|
| **baysg29** | 7.14 (10) | 6.88 (10) | 6.37 (10) | 7.36 (10) | 7.20 (10) | 7.26 (10) |
| **att48** | 7.52 (10) | 6.18 (10) | 6.47 (10) | 7.36 (10) | 6.52 (10) | 6.23 (10) |
| **eil51** | 15.42 (20) | 12.36 (20) | 12.54 (20) | 15.18 (20) | 14.83 (20) | 14.57 (20) |
| **st70** | 8.16 (10) | 6.44 (10) | 6.12 (10) | 7.18 (10) | 6.38 (10) | 6.24 (10) |
| **eil76** | 7.66 (10) | 6.85 (10) | 7.14 (10) | 7.84 (10) | 6.83 (10) | 6.57 (10) |
| **Kroa100** | 15.16 (20) | 13.36 (20) | 12.58 (20) | 14.94 (20) | 12.44 (20) | 12.21 (20) |
| **Kroa200** | 26.72 (30) | 20.82 (30) | 18.48 (30) | 21.76 (30) | 18.78 (30) | 18.15 (30) |
| **lin318** | 27.44 (30) | 23.13 (30) | 21.64 (30) | 23.95 (30) | 18.33 (30) | 18.68 (30) |

The Table 4 shows the details of average number of ant's usage for punished dynamic ants. The MRTS variant selects 60-70% and MTS, MeTS selects in the range of 50-60% of ants in the system for additional reinforcement. On comparison of Table 3 and Table 4, it can be observed that punishment mechanism selects lesser number of ants compared to non-punished mechanism.

The punishment mechanism demonstrate its ability in restricting the search in promising region by making use of optimal number of ants for additional reinforcement purpose. We continued the investigation on selection mechanism of ants for additional reinforcement by plotting the Box and Whisker graph, which provides the details of ant's selection distribution.

**Table 4: Average number of ants selected for second time updation in Punished Dynamic Ants**

| Data Sets | DEAMR | DEAM | DEA Med | DRAMR | DRAM | DRAMed |
|-----------|-------|------|---------|-------|------|--------|
| **baysg29** | 6.58 | 5.14 | 5.07 | 6.53 | 5.72 | 5.53 |
|  | (10) | (10) | (10) | (10) | (10) | (10) |
| **att48** | 6.15 | 5.03 | 5.01 | 5.26 | 5.18 | 5.23 |
|  | (10) | (10) | (10) | (10) | (10) | (10) |
| **eil51** | 13.84 | 10.48 | 10.36 | 14.26 | 12.20 | 11.78 |
|  | (20) | (20) | (20) | (20) | (20) | (20) |
| **st70** | 6.87 | 5.18 | 5.55 | 6.56 | 5.44 | 5.21 |
|  | (10) | (10) | (10) | (10) | (10) | (10) |
| **eil76** | 7.24 | 5.45 | 5.52 | 5.70 | 5.37 | 5.11 |
|  | (10) | (10) | (10) | (10) | (10) | (10) |
| **Kroa100** | 13.76 | 10.46 | 10.62 | 13.56 | 10.18 | 10.05 |
|  | (20) | (20) | (20) | (20) | (20) | (20) |
| **Kroa200** | 22.68 | 16.16 | 15.68 | 17.44 | 15.84 | 15.28 |
|  | (30) | (30) | (30) | (30) | (30) | (30) |
| **lin318** | 22.94 | 17.74 | 17.32 | 21.44 | 15.38 | 15.47 |
|  | (30) | (30) | (30) | (30) | (30) | (30) |

The Fig 1 and Fig 2 shows spread in the distribution of selected ants during algorithm execution for st70 and kroa100 datasets. We made some interesting observation and report it in more general conclusive manner. The Inter Quartile Range (IQR) was comparatively smaller for MRTS and skewed towards upper whisker. The Fig 1a and Fig 1b shows that DEAMR and DRAMR algorithms have 7 and 6 ants respectively at 50% (median) observation indicating that, most of the time algorithm selects larger number of ants. However, a better spread is observed in MTS and MeTS variants of algorithm and the observed median value is in the range of 5 to 6 ants. Another interesting observation is that, MRTS selects highest number of ants compared to MeTS and MTS.

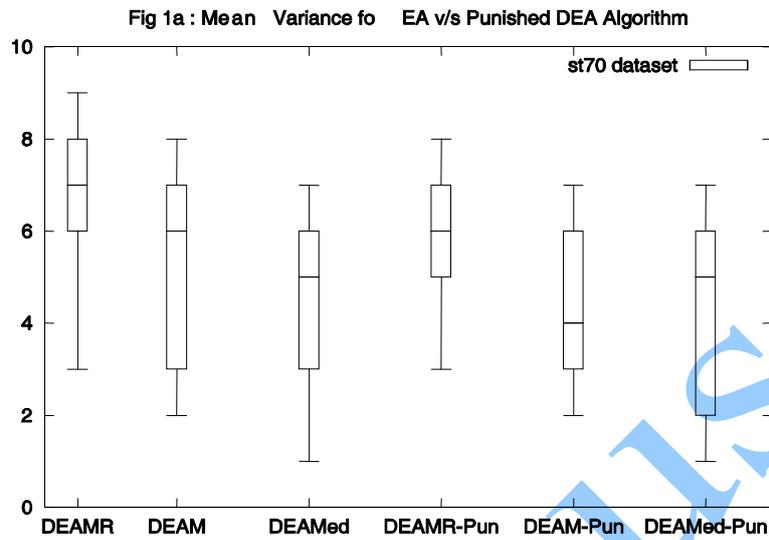

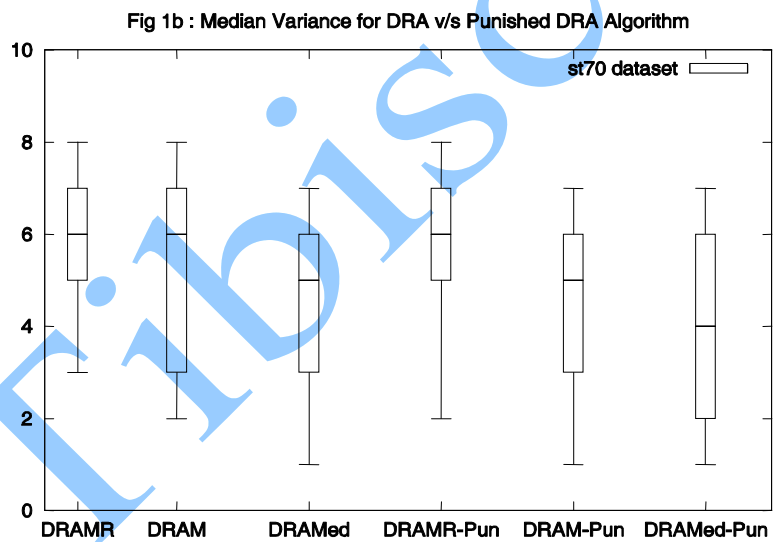

**Figure 1: Box and Whisker plot for distribution of selected ants
for additional reinforcement**

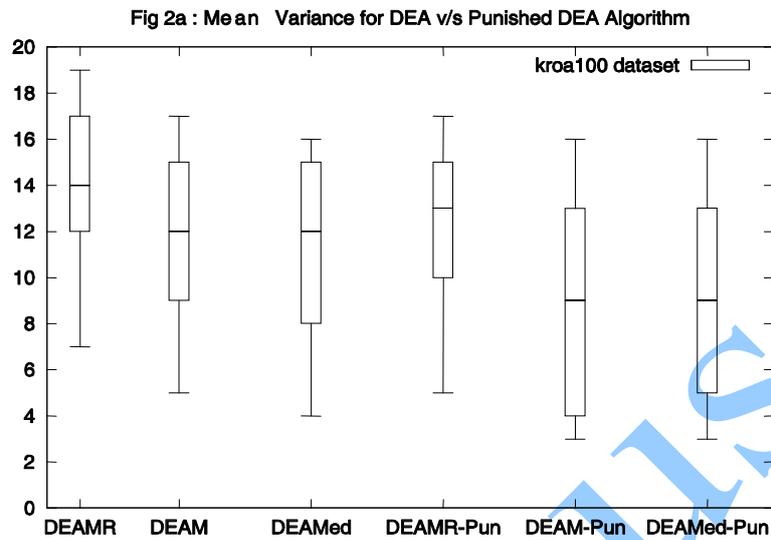

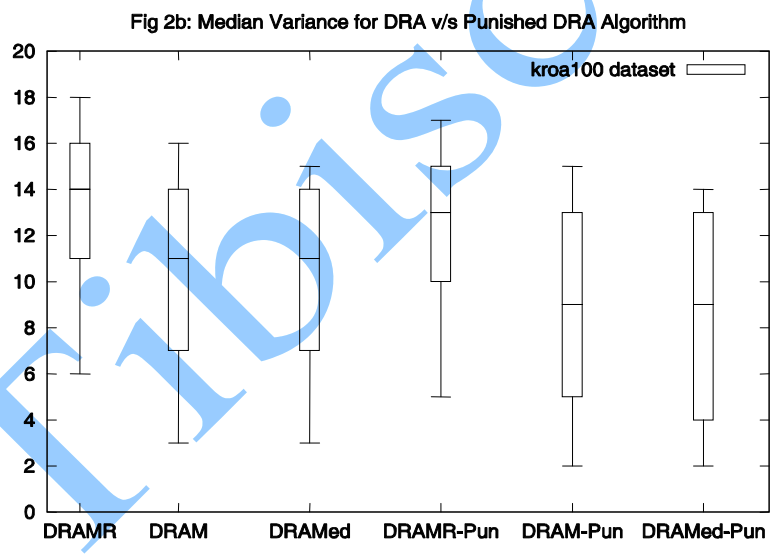

**Figure 2: Box and Whisker plot for distribution of selected ants for additional reinforcement**

## Conclusion

In this work, we have incorporated a classification mechanism that will classify the ants based on performance and learning mechanism that will impart knowledge to the ants for secondary updation. The classification mechanism strikes the balance between exploitation and exploration using statistical tools. The notable feature of proposed work is that, search carried in search space is dynamic in nature and number of elite path varies across the iterations. We extended the work by incorporating the punishment mechanism that has been more effective in restricting the search in the promising region of the search space and guiding the ants for looking optimal solution. We have provided a tight analysis of extended version of ACO and proved that in most variants obtained solutions are better than the optimal solution mentioned in the literature. Our future research concentrates on exploitation of regions near the obtained best solutions. The central idea is to train the ants such that nearby performances are updated with same quantity of pheromone trial with a hope that global best solutions are present near the local best solution.